\documentclass[sigconf]{acmart}
\AtBeginDocument{%
  \providecommand\BibTeX{{%
    \normalfont B\kern-0.5em{\scshape i\kern-0.25em b}\kern-0.8em\TeX}}}

\setcopyright{rightsretained} 
\copyrightyear{2023} 
\acmYear{2023} 
\acmConference{SIGGRAPH '23 Posters}{August 06-10, 2023}{Los Angeles, CA, USA}\acmBooktitle{Special Interest Group on Computer Graphics and Interactive Techniques Conference Posters (SIGGRAPH '23 Posters), August 06-10, 2023}\acmDOI{10.1145/3588028.3603653}
\acmISBN{979-8-4007-0152-8/23/08}

\begin{document}

\title{Reverse Projection: Real-Time Local Space Texture Mapping}

\author{Adrian Xuan Wei Lim}
\affiliation{%
  \institution{Roblox}
   \country{USA}
}
\email{xlim@roblox.com}

\author{Lynnette Hui Xian Ng}
\affiliation{%
  \institution{CMU}
   \country{USA}
}
\email{lynnetteng@cmu.edu}

\author{Conor Griffin}
\affiliation{%
  \institution{Roblox}
   \country{USA}
}
\email{cgriffin@roblox.com}

\author{Nicholas Kyger}
\affiliation{%
  \institution{Roblox}
   \country{USA}
}
\email{nkyger@roblox.com}

\author{Faraz Baghernezhad}
\affiliation{%
  \institution{Roblox}
   \country{USA}
}
\email{fbaghernezhad@roblox.com}

\renewcommand{\shortauthors}{Lim et al.}

\begin{abstract}
We present Reverse Projection, a novel projective texture mapping technique for painting a decal directly to the texture of a 3D object. Designed to be used in games, this technique works in real-time. By using projection techniques that are computed in local space textures and outward-looking, users using low-end android devices to high-end gaming desktops are able to enjoy the personalization of their assets. We believe our proposed pipeline is a step in improving the speed and versatility of model painting.
\end{abstract}


\begin{teaserfigure}
  \includegraphics[width=\textwidth]{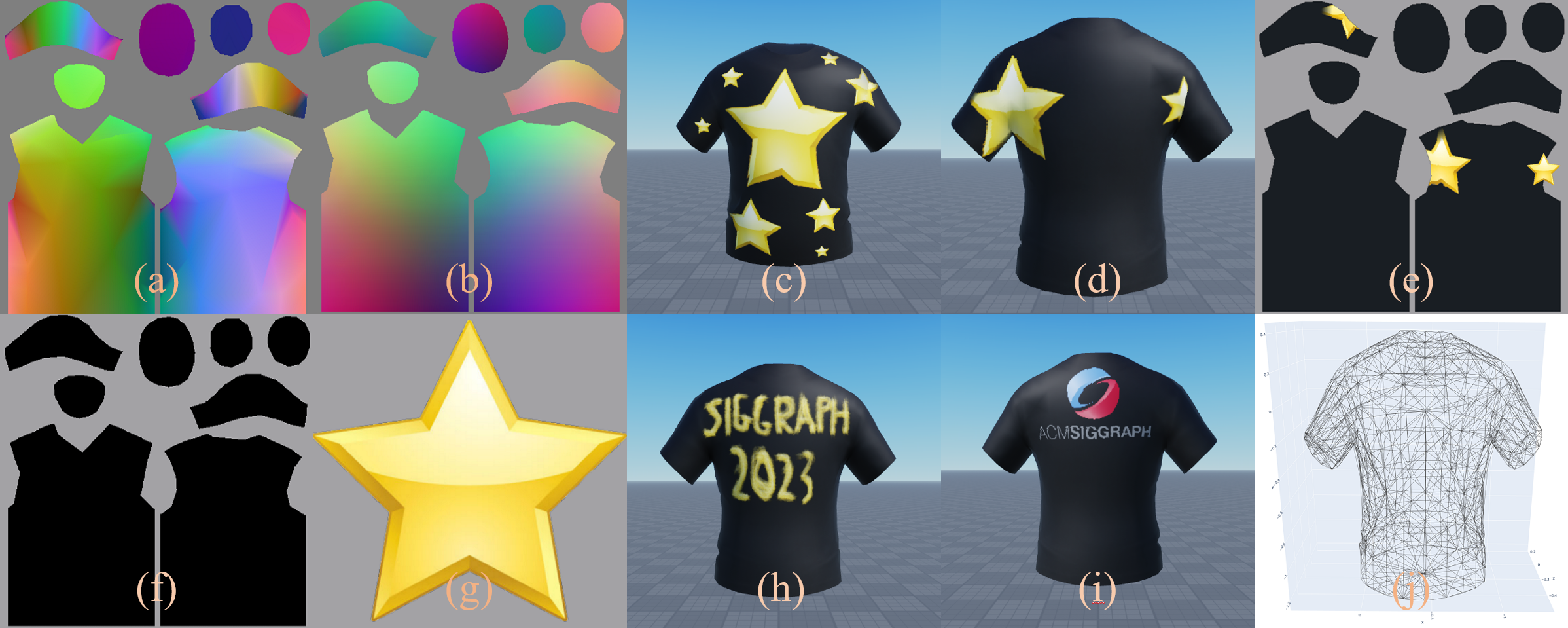}
  \caption{(a) Local Space Positional Map. (b) Local Space Normal Map. (c) Stamping with different projector transform (d) Stamping vs Projection Result. (e) Stamping vs Projection in Texture. (f) Target Texture. (g) Decal Texture. (h) SIGGRAPH Hand Painting using proposed technique. (i) SIGGRAPH Logo as Decal (j) Model Mesh.}
  \label{fig:teaser}
\end{teaserfigure}


\maketitle

\section{Introduction}
Reverse projection is a technique to project and color the surface of a 3D model. By updating the texture map of the 3D model from the model perspective, this provides the ability to customize and personalize the model's look and feel. This customization is especially important in games to design avatars and objects that fit within an overall theme.

Common modes of painting decals on 3D surface include ray tracing \cite{liktor2008ray}, uv-map unwrapping \cite{10.1145/364338.364404}, or forward projection mapping \cite{10.1145/2858036.2858329}.
Ray tracing identifies intersections between the decal and surface through shooting rays from the eye position at each pixel's center \cite{liktor2008ray}. Due to depth buffer quantization, the accuracy degrades.
UV-map unwrapping seeks a good surface parameterization that minimizes distortion for the texture mapping process \cite{10.1145/364338.364404}. However, this method does not account for edge detection as a uv-space of the 3D model is not continuous to the surface space for the texture map. 
Lastly, forward projective mapping involves projecting light bundles to the target object. This technique relies on extensive tracking of projection features to correctly align the decal to the texture \cite{10.1145/2858036.2858329}.

Both ray tracing and uv-map unwrapping have their associated compute costs, while forward projection mapping does not directly draw into the target's texture. For model painting to be performed on edge devices and within real-time systems like games, we want to reduce compute cost by direct painting on the target texture.


In our approach, we reframe the projection problem by making individual pixels on a target texture look outwards and singularly determine the requirement and positionality of projection. We develop a pipeline that harnesses per-pixel operations to perform a projection routine for 3D model painting. 



\section{Methodology}
\begin{figure}[htbp]
\centering
\includegraphics[width=0.5\textwidth]{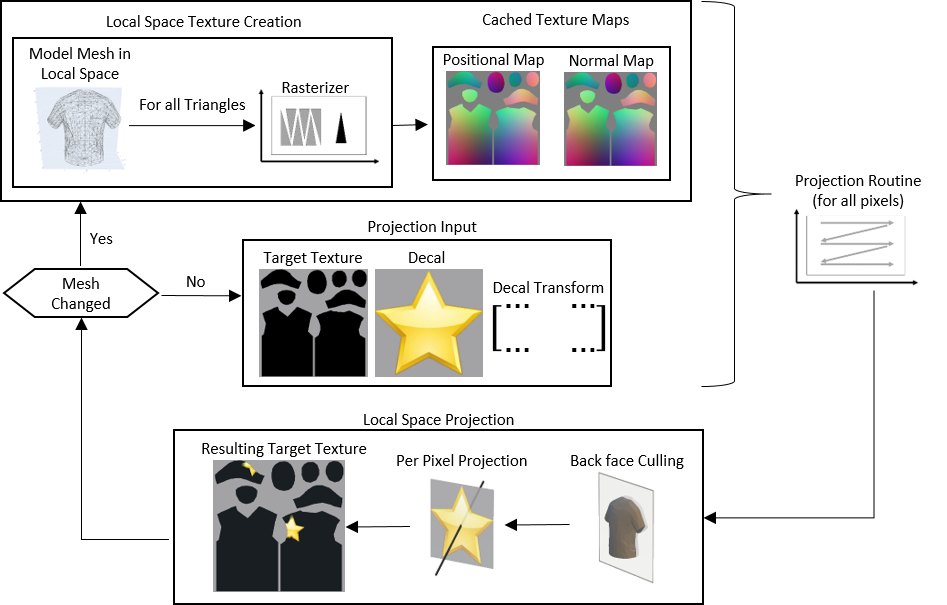}
\caption{Reverse Projection Framework}
\label{fig:Framework}
\end{figure}

\autoref{fig:Framework} illustrates the overview of the processing pipeline for our Reverse Projection framework. To paint the model, we convert the model mesh into a local space texture, thus reducing the required mathematical operations into texture interactions, allowing for real-time projection mapping.

\paragraph{Local Space Texture Generation.} For the target model, we obtain the triangle mesh in local space. We apply a rasterization function that iterates all the triangles in the mesh and linearly interpolates the position and normal information for each triangle. The function produces an overall positional and normal map. The positional map is a point cloud texture. The normal map requires 3-axis of information ($x,y,z$), as compared to 2 from a traditional tangent-space normal map. This step is skipped if the mesh is not changed.

\paragraph{Projection Input}
Given a target texture and a decal (texture to be painted), we apply a decal transform operation. This transform matrix provides the position, orientation and scale of the decal with respect to the local space of the model mesh. The output feeds into a projection routine to perform Local Space Projection per pixel.

\paragraph{Local Space Projection}
Back face culling is performed to ignore surfaces that are faced away from the user, optimizing the computational cost. A series of dot products are performed between the texture and decal to determine which pixels to keep. For these pixels, a ray cast operation is performed to determine whether the position of the texture touches the decal. The ray intersection test also returns the position of the decal, given which we can copy the color to the texture, producing our painted model.


\section{Results and Discussion}
\autoref{fig:teaser} shows images of milestone output steps. This pipeline demonstrates a method to project a decal in a computationally inexpensive method for direct painting of textures.

We use a Dell Precision 5570 (Intel i9-12900H with 64gb of DDR4 4800Mhz) for our evaluation. 
We implemented purely single-threaded CPU functions in our pipeline, to account for cases where some devices do not have threading abilities. This includes the rasterization and projective mapping functions, which can be done in compute shaders or multiple threads for efficiency. With this setup, the \textit{Local Space Texture Generation} step completed in $0.0033\pm0.003$s which requires iterating all triangles in the mesh. Each triangle on the texture surface is unique, and the total triangle count should not exceed (width*height) of the texture.


Due to the uniqueness of each Local Space Texture coordinate, the upper bound of processing the positional/normal map will also be (width*height). In total, the processing upper bound of our pipeline is O(2*width*height).
Our \textit{Local Space Projection} runs completed in $0.0033\pm0.003$s. 
Thus, our method is optimal because the compute complexity is never larger than that required to update the target texture.


Limitations of this method include: all triangles must uniquely map to an area on the texture to provide information for rendering. Therefore the mesh triangle count cannot exceed local space texture size. The target texture width/height must match the local space texture size. 
Future work calls for improvements to painting techniques like broad phase culling for complex non-parametric models \cite{10.1145/566654.566649}.

 
In conclusion, our method successfully produces a painted 3D model by using reverse projection texture mapping in local space. Our pipeline optimizes the compute complexity and time such that the technique can be implemented on devices with different compute ranges. 


\begin{acks}
Much thanks to Alexander Ehrath, Ebor Folkertsma, Morgan McGuire from Roblox and the rest of the amazing folks at Roblox and CMU.
\end{acks}

\bibliographystyle{ACM-Reference-Format}
\bibliography{main}

\end{document}